\documentclass[conference]{IEEEtran}
\IEEEoverridecommandlockouts                              
\usepackage[bookmarks=false]{hyperref}
\usepackage{blindtext, graphicx}
\usepackage{cite}
\usepackage{graphicx}
\usepackage{subfigure}
\usepackage{fixltx2e}
\usepackage{csquotes}
\usepackage{caption}
\usepackage{amssymb}
\usepackage{amsmath}
\usepackage{siunitx}
\usepackage[font={small}]{caption}	% Reduce the caption font size
\usepackage{fancyhdr}

\graphicspath{{Figures/}}
\newcommand{\fig}[1]{Fig.~\ref{#1}}
\newcommand{\eq}[1]{(\ref{#1})}

\usepackage[referable]{threeparttablex}

\DeclareCaptionLabelSeparator{periodspace}{.\quad}
\usepackage{watermark}
\watermark{\hspace{-0.3in} {SS13: Mixed-signal Circuits for Machine Learning and Edge-AI} \hspace{2.0in}}

\begin{document}
\title{Metaplasticity in Multistate Memristor Synaptic Networks}
\author{\IEEEauthorblockN{Fatima Tuz Zohora$^{\S}$, Abdullah M. Zyarah$^{\dagger}$, Nicholas Soures$^{\dagger}$, Dhireesha Kudithipudi$^{\S}$}
\IEEEauthorblockA{Neuromorphic AI Lab, University of Texas at San Antonio$^{\S}$ \\ Rochester Institute of Technology$^{\dagger}$}}
\IEEEoverridecommandlockouts
\IEEEpubid{\makebox[\columnwidth]{~XXXX-XXXX~\copyright~2020 
IEEE \hfill} \hspace{\columnsep}\makebox[\columnwidth]{ }}
\maketitle

%%%%%%%%%%%%%%%%%%%%%%%%%%%%%%%%%%%%%%%%%%%%%%%%%%%%%%%%%%%%%%%%%%%%%%%%
\begin{abstract}
%High retention of learned information and high reception to new information over time is a highly desired characteristic of synaptic networks.Plasticity mechanisms that enable a synapse to change its response to activity i.e. mataplasticity can help synapses achieve this behavior.
Recent studies have shown that metaplastic synapses can retain information longer than simple binary synapses and are beneficial for continual learning. 
In this paper, we explore the multistate metaplastic synapse characteristics in the context of high retention and reception of information. 
Inherent behavior of a memristor emulating the multistate synapse is employed to capture the metaplastic behavior. An integrated neural network study for learning and memory retention is performed by integrating the synapse in a $5\times3$ crossbar at the circuit level and $128\times128$ network at the architectural level. An on-device training circuitry ensures the dynamic learning in the network. In the $128\times128$ network, it is observed that the number of input patterns the multistate synapse can classify is $\simeq$ 2.1x that of a simple binary synapse model, at a mean accuracy of $\geq$ 75\% .  
\end{abstract}


% \let\thefootnote\relax\footnote{This material is based on research sponsored by AirForce Research Laboratory under agreement number FA8750-16-1-0108. The U.S. Government is authorized to reproduce and distribute reprints for Governmental purposes notwithstanding any copyright notation thereon. \\
% The views and conclusions contained herein are those of the authors and should not be interpreted as necessarily representing the official policies or endorsements, either expressed or implied, of AirForce Research Laboratory or the U.S. Government.}

%%%%%%%%%%%%%%%%%%%%%%%%%%%%%%%%%%%%%%%%%%%%%%%%%%%%%%%%%%%%%%%%%%%%%%%%

\section{Introduction }

%\item Why is it important to design metaplastic memristor synapses? What is the motivation?\\--DK
%\textcolor{blue}{ Highlevel statement on benefits of plasticity in NN.}
%\item what is metaplasticity\\ 

%>>>> AMZ: Did not understand the second part of the first statement
Neural plasticity in the brain is the ability to learn and adapt to intrinsic or extrinsic stimuli by reorganizing the morphology, functions, or connectivity of its constituent synapses and neurons. Synaptic plasticity is a complex dynamic process that modulates and regulates network dynamics depending on external activity over multiple timescales. 
Metaplasticity refers to \textit{plasticity of the plasticity} of synapses \cite{abraham1996metaplasticity}. A metaplastic synaptic network enables a synapse to tune its level of plasticity depending on the pre-synaptic activity. This property is deemed crucial for high memory retention and learning capability in a synaptic network\cite{fusi2005cascade}.
%\item serial metaplastic synapse and its advantages\\
%does not necessarily manifest as a change in synaptic efficacy. A metaplastic synaptic network captures the intelligence to modify its responsiveness to new activity based on the previous activity of the network.
%\textbf{It is shown that simple binary synapses are well-suited to highly sparse neuronal activity, but when the activity becomes less sparse, the interference between multiple stimuli poses challenge}. 
It is shown that simple binary synapses show high memory retention when the imposed activity is highly sparse. However, for moderately sparse neuronal activity, the interference between multiple stimuli can pose a challenge to achieve high memory retention and learning. Since binary synapses cannot concurrently learn new activity and retain knowledge of past activity, the synapse memory lifetime drops significantly \cite{leibold2007sparseness}. To solve this issue, Fusi et al. \cite{fusi2005cascade} proposed a cascade model of synapse, in which synapses with binary efficacy have multiple metastates. Synapses exhibit varying degree of plasticity depending on their metaplastic state. This property enables a network of such synapses to retain knowledge of past activity and facilitate high plasticity to learn new activity. 
%\textbf{However, the maximal achievable memory lifetime in cascade synapses is orders below that of binary synapses for highly sparse activity, and it degrades with increasing number of metastates.} 
While the cascade synapse outperforms a simple binary synapse in response to moderately sparse activity, its memory retention for highly sparse activity is orders below that of a simple binary synapse. In \cite{leibold2007sparseness}, Leibold et al. proposed a variant of metaplastic synapse model, in which the metastates are serially connected and the probability to transit from one state to another is equally likely. This serial synaptic metaplasticity model, also referred to as multistate synapse shows less degradation in memory lifetime for highly sparse activity and outperforms the cascade model in memory capacity \cite{leibold2007sparseness}.
%although the latter requires fewer number of metastates for a given network size\cite{leibold2007sparseness}.
In this paper, we focus on the multistate synaptic model. Previous research on metaplasticity focused on physical metaplastic behavior in memristor devices \cite{zhu2017emulation,wu2018full,kim2017nanogenerator}. Most of the prior literature is concentrated on device level analysis considering only continuous synaptic efficacy with no network level realization. However, incorporating metaplastic synapses in a crossbar architecture can lead to compact and powerful neuromorphic architecture capable of high memory retention. Since edge devices encounter large amounts of streaming data, such architecture can immensely benefit their overall performance.

One of the early realizations of the binary metaplastic synapse was proposed by \cite{leibold2007sparseness}. Since this model can retain previously learned information and maintain response to new information simultaneously, such a synaptic model can better capture all the information learned throughout its lifetime. Hence, it shows better resilience against catastrophic forgetting compared to binary synapses. In this research, we study this synaptic model at-scale in memristive neural accelerators. The main contributions of this paper are as follows:
\begin{itemize}
\item to emulate binary metaplastic synapses by exploiting inherent device properties of a memristor. 
\item to demonstrate the efficacy of metaplastic synapse in a $5 \times 3$ crossbar circuit architecture with on-device learning capability.
\item to compare the performance of binary vs. metaplastic synapse in a two layer neural network emulating hardware constraints.
%\item HIgh level simulation of a network of multistate synapses considering hardware constraints and comparison with its binary counterpart.
\end{itemize}
%We analyze the model behavior to show that the multistate synapse can retain information learned throughout its lifetime with higher mean accuracy than binary synapse. 
%>>>> AMZ: I do not believe this is a valid statement!! "we design a metaplastic synapse with memristor " Back the my comment in the abstract 
%Furthermore, we emulate the behavior of a metaplastic synapse with memristor exploiting its inherent device property and demonstrate its efficacy in a crossbar accelerator architecture with on-device learning capability. We emulate the hardware metaplastic synaptic model in a high level simulation to compare its performance with its binary counterpart. 
%We observe that the number of input patterns that a multistate synapse model can detect with mean accuracy $\geq$ 75\% is $\simeq$ 2.1x times 
%>>>> AMZ: 2.2 times more than that ....
% FTZ: 2.2 times more sounds 3.2 times. so i omitted more.that of a binary model for a $128\times128$ network. 

%Hence, such synapses may prove to be effective in more complex tasks as well to tackle catastrophic forgetting. 

%\item Goal of this research and contributions\\

%%%%%%%%%%%%%%%%%%%%%%%%%%%%%%%%%%%%%%%%%%%%%%%%%%%%%%%%%%%%%%%%%%%%%%%%
\section{Metaplastic Synaptic Network Model}

 The multistate synapse is a relatively simple model where metaplasticity is modeled by serially connected metastates. The probability to transit from one state to the other is equal. \fig{fig:syn} shows the metastates of the multistate synapse and their inter-transitions. The red and blue bubbles represent synaptic metastates with efficacy 1 and 0 respectively. The arrows show the transition direction, the red  arrows correspond to potentiation and the blue arrows represent depression. As shown in \fig{fig:syn}, the synapse changes its efficacy only when it is in metalevel ($\eta$) 0; in all other cases it only changes the metalevel retaining its efficacy. Multistate model with n metalevels can exhibit (2n-1) forgetting timescales which helps it to retain knowledge of past activity \cite{leibold2007sparseness}.
 
In \cite{ben2007long} and \cite{leibold2007sparseness}, the authors investigate memory lifetime by imposing a specific pattern of activity to the network and observing how long the network can recollect the learned information. It is shown that complex synapses with metaplasticity can retain information longer than simple binary synapses when the neuron activity becomes less sparse. In this work, we explore how metaplasticity affects the accuracy of a synaptic network to detect all the patterns learned throughout its lifetime and its  capability to learn new activity. We consider a simple feed forward network where $N_{in}$ input neurons are connected to $N_{out}$ output neurons through a network of sparse synapses. Random input patterns and corresponding output patterns of activity \textit{f} (\textit{f}\% of bits are high) are generated and applied to a network with connectivity \textit{C}, i.e. \textit{C}\% of the input and output neurons are connected to each other. Initially, the connected synapses have random efficacy and they are at their most plastic state. Similar to \cite{leibold2007sparseness}, we use McCullogh-Pitts neuron model at the output nodes. This neuron detects activity if the incoming signal is greater than its threshold, which is set based on the average input to an output neuron in the randomly initialized network. In a network with connectivity \textit{C} and input activity \textit{f}, the threshold is equal to $N_{in}\textit{Cf}/2$. We use an error based learning rule to train the network, where $error= (y-y_{n})$  ($y$ is the ground truth label and $y_{n}$ is the network output) and only the synapses with active presynaptic inputs are updated. A synapse is potentiated for positive error and depressed for negative error. Using this set up, we train $128\times128$ networks (f=25\%, C=25\%) of simple binary and multistate synapses. We also train a similar sized network with gradient descent (GD) in which the synaptic weights are thresholded for computation. Two types of accuracy are tracked in the networks, (1) accuracy to detect the most recent input to evaluate learning capability and (2) the mean accuracy across all the patterns encountered, to evaluate the network's resilience against catastrophic forgetting. 

\begin{figure*}[h!tb]
\centering
\subfigure{\includegraphics[width=60mm, height=45mm]{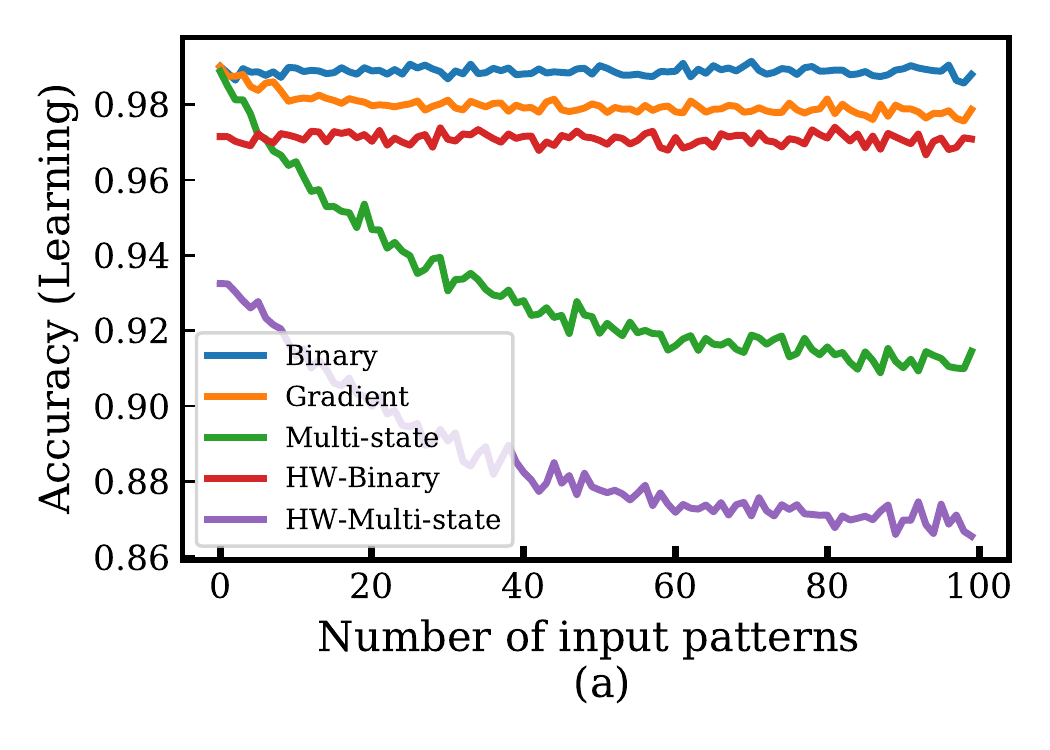}}
\subfigure{\includegraphics[width=60mm, height=45mm]{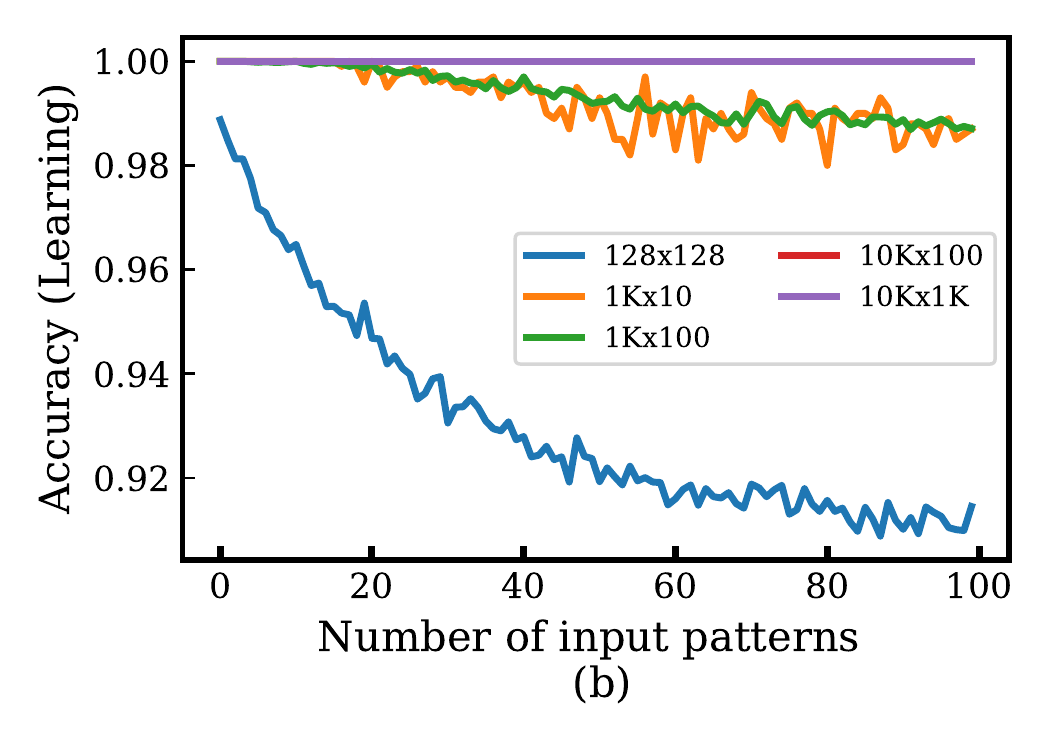}}
\subfigure{\includegraphics[width=60mm, height=45mm]{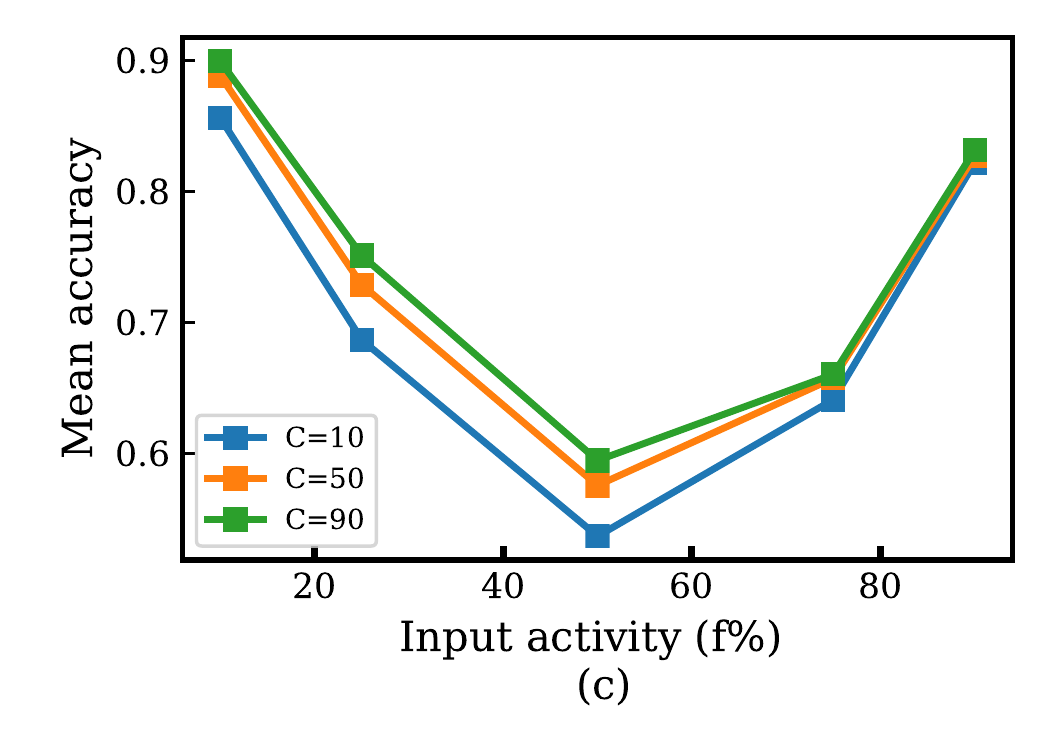}}
\subfigure{\includegraphics[width=60mm, height=45mm]{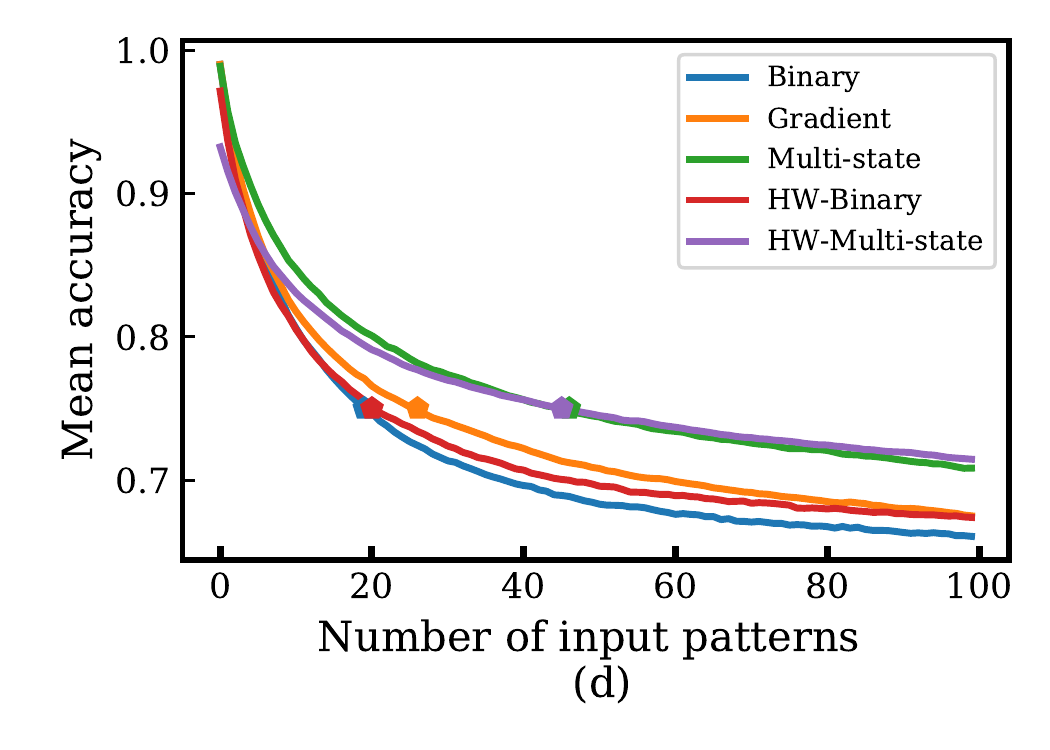}}
\subfigure{\includegraphics[width=60mm, height=45mm]{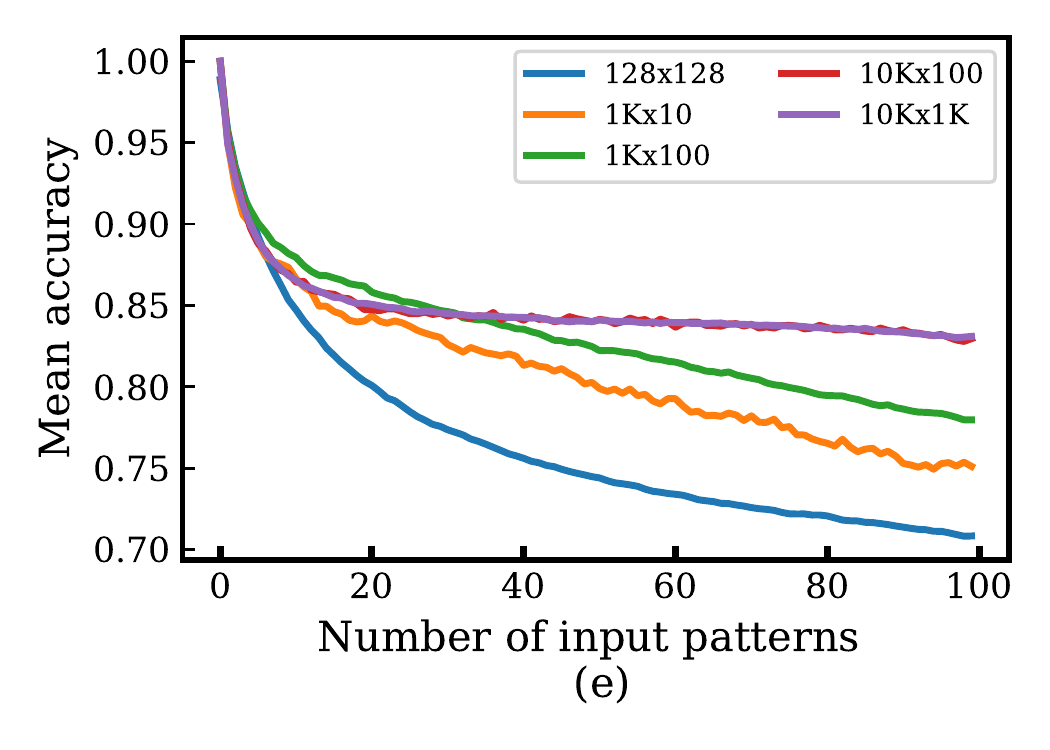}}
\subfigure{\includegraphics[width=60mm, height=45mm]{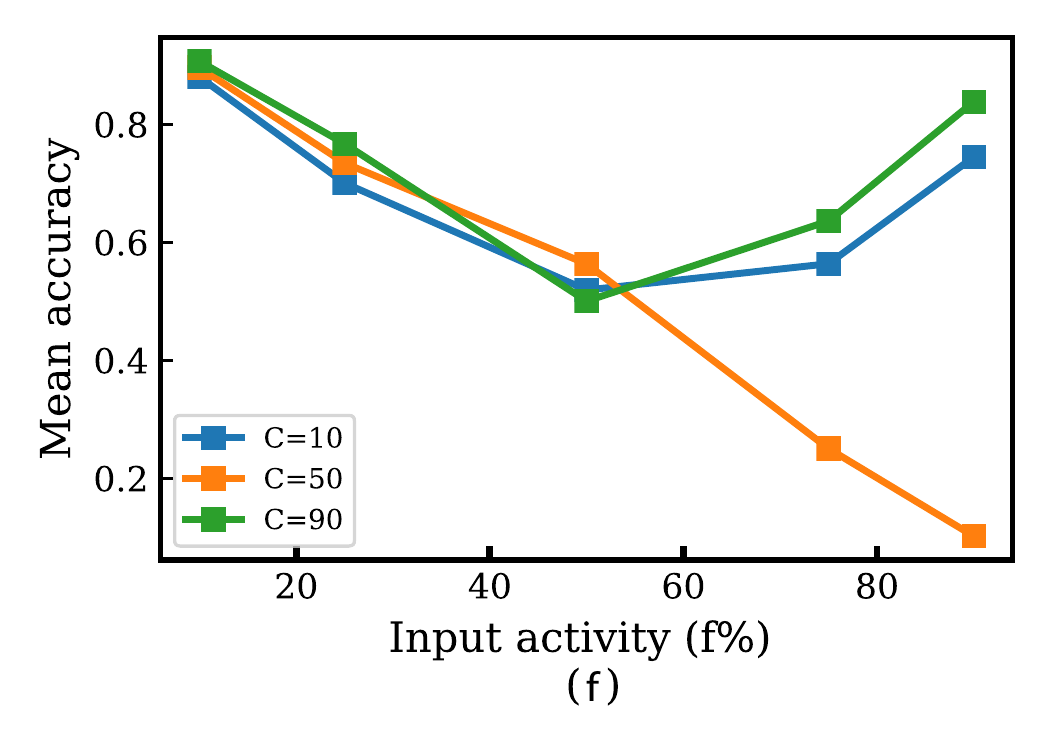}}

%>>>> AMZ: Learning and mean accuracy of $128\times128$ network developed using Binary and Multistate. Here, H/W-Binary and H/W-Multistate refer to the same network emulated in hardware while considering all devices non-idealities . I'm still not convinced with my sentence! still try to find a better way of saying this ... You can help! Assuming you like what I suggested. 
% not all non idealities! Some right? I do just without all its right i guess
\caption{In (a) \& (d): Learning and mean accuracy of $128\times128$ network developed using binary and multistate synapses. Here, H/W-Binary and H/W-Multistate refer to the network designed with hybrid CMOS/memristor circuitry in Cadence, while considering device non-idealities. Gradient shows the accuracy for network trained with gradient descent. In (b) \& (e): 
%>>>>> AMZ: suggestion: Learning and mean accuracy of multistate networks as a function of the network size.
Learning and mean accuracy of multistate networks as a function of the network size. In (c) \& (f): Effect of network connectivity \textit{C} and input activity \textit{f} on the mean accuracy after presenting 100 patterns to a $128\times128$ network of multistate synapses and its hardware emulation respectively.\vspace{-2mm}}
\label{metric}
\end{figure*}

\vspace{-2 mm}
In \fig{metric}(a) we see that the binary network outperforms both GD and multistate network in learning accuracy, the multistate network shows $\simeq$ 91\% accuracy after encountering 100 patterns whereas for the binary network it is $\simeq$ 99\%. However, in \fig{metric}(d) we see that the mean accuracy drops significanly slower in the multistate network than both GD and binary network. To compare the performance across networks we empirically set a threshold at 75\% for the mean accuracy and observe the number of imposed patterns after which the mean accuracy goes below it. From \fig{metric}(d) we see that the mean accuracy goes below the threshold after 20 patterns for the binary network whereas in multistate network it is 45 patterns.  The loss in learning accuracy is significantly lower than the gain in mean accuracy.  We also observe in \fig{metric}(b) and (e) that with increasing network size the multistate network shows less degradation in learning accuracy and even higher mean accuracy as its memory capacity grows with size. We further investigate the effect of network connectivity \textit{C} and input activity \textit{f} on the mean accuracy of a multistate network (\fig{metric}(c)). We notice a drop in the performance around \textit{f}=50\% due to rise in conflicting patterns, but it retains high performance for high and low connectivity. Overall, the simulation results indicate that the multistate synaptic model shows better accuracy in detecting patterns learned over lifetime accompanied by a degradation in learning ability compared to binary and it is particularly suitable for modeling large network of synapses.

\section{Modeling Multistate Synapses with Memristors}
\begin{figure}
    \centering
    \includegraphics[width=0.9\linewidth]{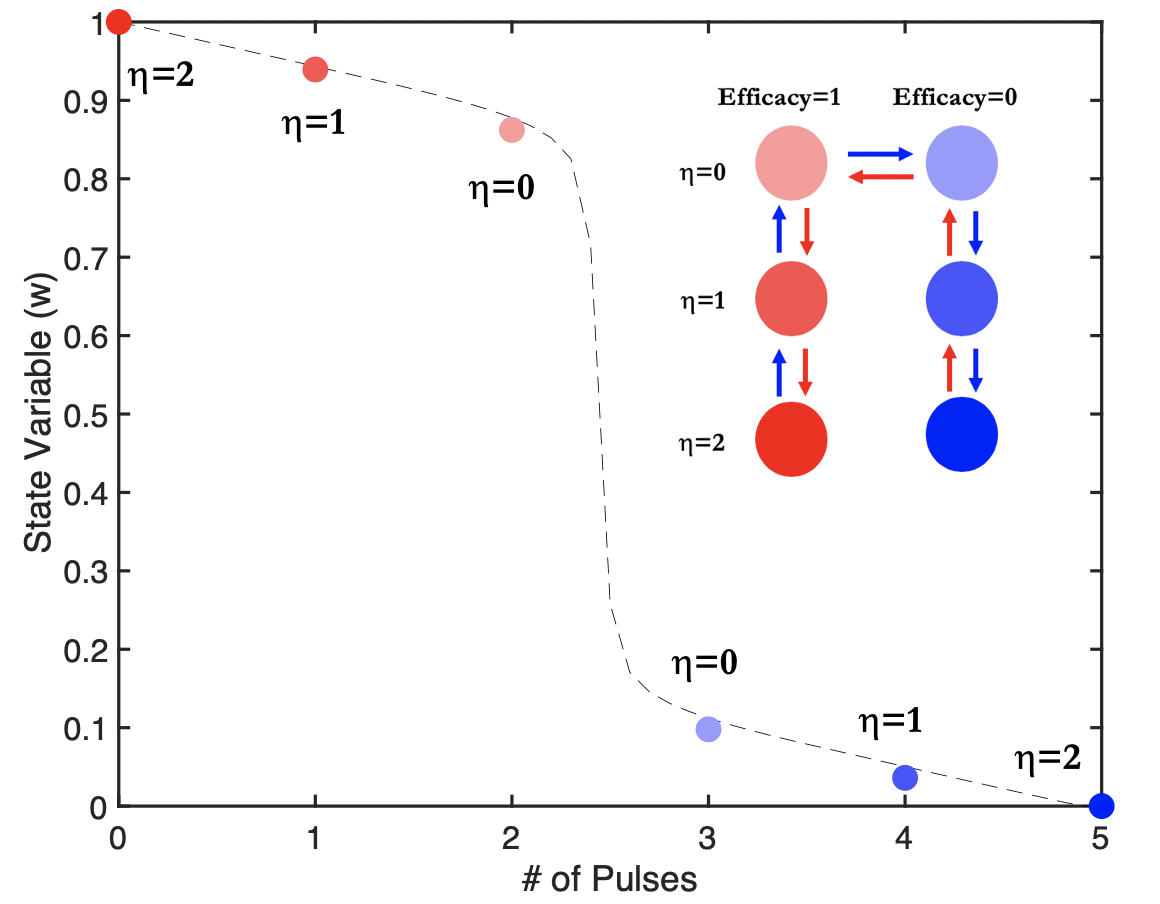}
    \caption{Representation of the multistate metaplastic synapse model mapped to a physical memristor device behavior captured from \cite{jiang2017rram}.\vspace{-6mm}}
    \label{fig:syn}
\end{figure}

%\begin{itemize}
%\item  how can it be abstracted with memristor behavior\\
In this work, we leverage device characteristics of memristor device to realize a multistate synapse. As presented in \cite{jiang2017rram}, the device under consideration shows gradual change in conductance during RESET. For modeling we assume this behavior during SET operation as well. To emulate the metastates, the memristor was trained with 15$\mu$s pulses of 1.2V to potentiate or depress from one state to another. In this process, we get three states with \textit{high} and \textit{low} conductance  which can represent different metastates. The correlation between the metastates and the memristor state variable ($w$) which is proportional to its conductance is shown in \fig{fig:syn}. In the ideal multistate model, change in metalevel incurs no change in synaptic efficacy. However, in the hardware emulation the conductance of the memristor varies across metastates. The device has to be programmed to ensure that the difference in conductance between the \textit{high} and \textit{low} efficacy states is substantial. In the modeled memristor, the lowest and highest resistive states were set to be $100k\Omega$ and $10M\Omega$ respectively and the ratio of conductance between high and low efficacy state at metalevel zero is $\simeq$ 4.5.

% \\Simulation-Fatima 
%------------ Fatima ---------------
%>>>> AMZ: Suggestion: Representation of the multistate metaplastic synapse model   . 

%--------- Abdullah ------------
A modified Verilog-A memristor model proposed by~\cite{kvatinsky2015vteam} is employed to model the memristor. The device conductance changes as a function of the state variable, $w$, which is described in~\eq{mem_eq} and~\eq{mem_eq2}\footnote{$k_{off}$, $k_{on}$, $\alpha_{on}$, and $\alpha_{off}$ are constants, and $v_{off}$ and $v_{on}$ are the memristor threshold voltage.}, where $D$ is the device thickness, and $G_{on}$ and $G_{off}$ define the memristor conductance limits.  Here, the memristor model is tightened with a modified Z-window function~\cite{zyarah2019neuromemrisitive} (see~\eq{mem_eq3})\footnote{$\tau$, $\delta$, and $p$ are constants to control the window function shape.}\vspace{-1mm}
 
\begin{equation}
G_{mem} = \frac{w}{D} \times G_{on} + (1 - \frac{w}{D}) \times G_{off}
\label{mem_eq}
\end{equation}
\begin{equation}
\frac{\Delta w}{\Delta t} = 
\begin{cases}
k_{off}.\Big(\frac{v(t)}{v_{off}} - 1\Big)^{\alpha_{off}}.f_{w}(w),&0 < v_{off} < v \\
0, &v_{on} < v< v_{off} \\
k_{on}.\Big(\frac{v(t)}{v_{on}} - 1\Big)^{\alpha_{on}}.f_{w}(w),&v <v_{on} < 0
\end{cases}
\label{mem_eq2}
\end{equation}
\begin{equation}
\label{mem_eq3}
    f_w = \frac{1-4(\frac{w}{D} - \delta)^2}{e^{\tau (\frac{w}{D} - \delta)^p}}
\end{equation}

To account for device variability, a random Gaussian noise with standard deviation of 25\% is induced for 100 cycles. The amount of noise considered is relatively higher than observed in actual devices \cite{hu2011geometry} to compensate for the suppression of variability due to the Z window function.
%This memristor model is simulated with random Gaussian noise for 100 cycles and the resulting mean device conductances for each metastate is used for higher level simulation. 

\section{Metaplastic System Level Design}
\begin{figure}
    \centering
    \includegraphics[width=0.9\linewidth]{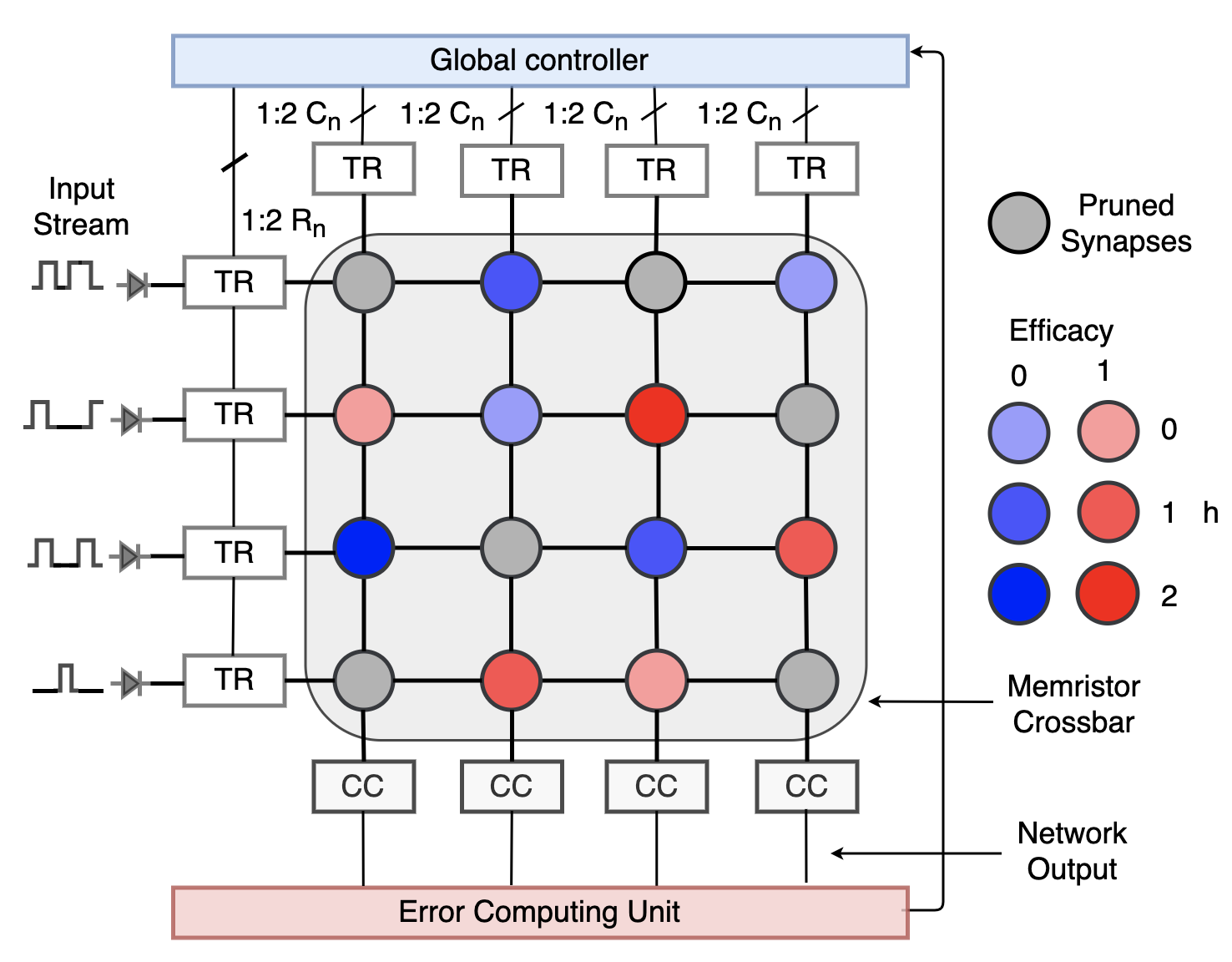}
    \caption{Sytem level architecture of the memristor metaplastic network, with low sparsity. As denoted, the memristors can be at different synaptic efficacies or pruned during learning. \textit{TR} represents the row and column training circuitry and \textit{CC} is the current comparator circuitry.\vspace{-3mm}}
    \label{fig:arch}
\end{figure}

\begin{figure}
    \centering
    
    \includegraphics[width=0.7\linewidth]{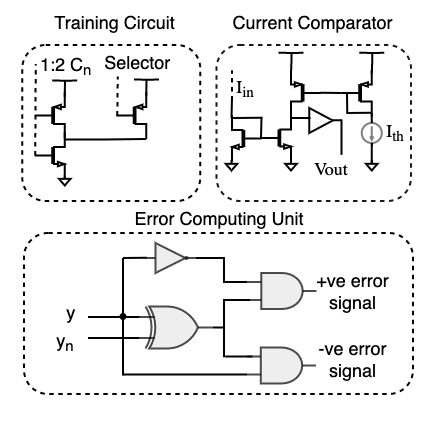}
    \caption{The training circuit, current comparator and the error computing unit for one crossbar column within the network. 
    %>>>>> AMZ: suggestion: you do not need to say what y and yn are in the caption 
   }
    \label{fig:tr}
\end{figure}

%>>>>> Fatima, would not be better or more clear if you say that the multistate synaptic behaviour has been explored or studied using two layers networks. Then, you can start talking about the system architecture.  
The multistate synaptic behavior is studied using the system architecture of a two layer neural network shown in \fig{fig:arch}. 
The sparse synaptic network is emulated by a crossbar consisting of the memristor model described in \fig{fig:syn}.
%>>>>>> long and unclear
The crossbar is initialized randomly maintaining connectivity $C$, setting the synapses with \textit{high} and \textit{low} efficacy to the most plastic metalevel ($\eta=0$). We model the sparse connectivity by  randomly setting crosspoints between the two neuronal 
%>>>>>>> What do you mean by unconnected? I do not know if it coveys the meaning you want. 
layers to the lowest conductivity which are left untrained. 
A current comparator is chosen to model the McCullogh-Pitts neuron and we exploit its varying input resistance to improve the network performance. 
%Input node to an output neuron consists of a diode connected transistor, which presents a variable input resistance to a synaptic column depending on its overall resistance. 
The input resistance of the neurons changes in the same direction as the column resistance which improves the networks ability to detect presynaptic activity and increases the mean accuracy.
Inference is carried out simultaneously for all the columns. Since the synaptic columns are not grounded, if the inputs are directly connected to the output neuron, voltage drop across the current comparator will push current back into the input nodes with inactive presynaptic input. To prevent this, the inputs are connected to the crossbar through diodes.   
%>>>>> The reader may not be aware to the fact that the voltage drop across the output neuron transistor may push the current back to the input node with zero voltage due to the potential difference. You need to elaborate more here I believe! 
The programming scheme described in section III is executed using Ziksa \cite{zyarah2017ziksa} to carry out the synaptic transitions. In this work, Ziksa is slightly modified by adding an extra transistor to the row trainer. This extra transistor holds the rows with inactive presynaptic input at $V_{DD}/2$
%>>>>> Suggestion: to reduce sneak current in crossbar during training.
to reduce sneak current in the crossbar during training.  The network is trained in two steps for the synapses to be potentiated and depressed, respectively. The training circuitry, current comparator and the error computing unit are shown in detail in \fig{fig:tr}.
%>>>>>>> THE FLOW IS MISSING!

%Since the most plastic synapses with low efficacy have significantly higher conductance than $G_{off}$, it is important to assess their contribution to the threshold current.   Considering these hardware constraints, a $128\times128$ crossbar is simulated (f=0.25 and S=0.25) for binary of multistate synapses with a Gaussian noise ($\sigma$ =0.25) to account for device variability. 

\section{Results and Analysis}
\begin{figure}
\centering
\includegraphics[width=0.9
\linewidth]{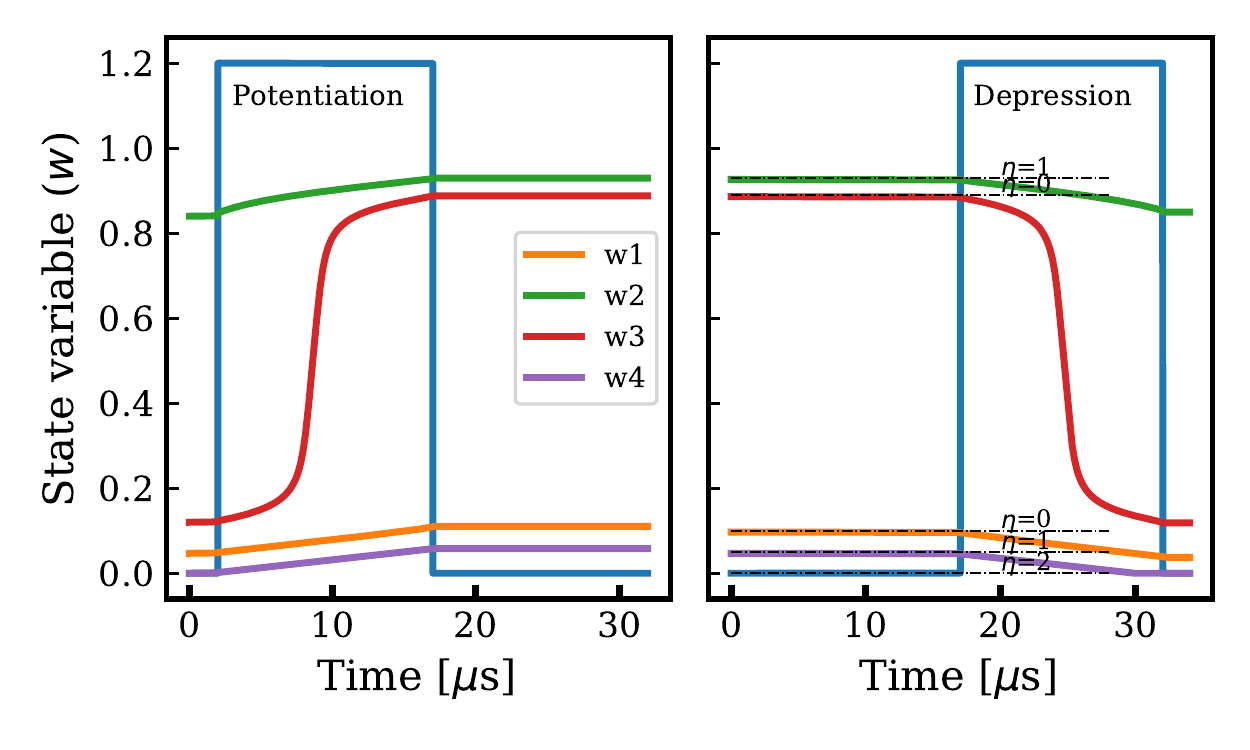}
\caption{Change in metaplastic synapses with potentiation and depression. The left panel shows the change in synapses with potentiation and the right panel shows the same for depression. The dotted lines in the right panel show the metalevels of the synapses. We see that only the synapses in metalevel 0 ($w_{2}$ in both panels) changes its efficacy level while the synapses in higher metalevels ($w_{1}$,$w_{3}$and $w_{4}$) only change their metalevels retaining the same efficacy level.\vspace{-6mm}}
\label{dual}
\end{figure}

%>>>>> Start with something like: In order to evaluate the proposed system architecture.... or The proposed system architecture is evaluated using ...
The proposed architecture is evaluated through high level simulation of $128\times128$ networks (\textit{f}=25\%, \textit{C}=25\%) of binary and multistate memristive synapses with random Gaussian noise ($\sigma$=0.25). The simulation is carried out in MATLAB with implication of  hardware constraints.
%>>>>>>> suggestion: complete the sentence as follows: show the learning and mean accuracy for hardware emulation for binary and multistate synapses while performing / doing / ... what ever task you doing 
\fig{metric}-(a) and (d) show the learning and mean accuracy for the hardware emulation of binary and multistate synapses while 100 random input patterns {similar to \cite{ben2007long}} with activity \textit{f} are imposed on the network. We see that the mean accuracy for
binary network drops below 75\% after presenting 22 input patterns whereas for the multistate network this accuracy drop is observed after learning 47 patterns. Similar to the analysis in section II, the high mean accuracy comes at a cost of drop in the learning capability. The drop in the learning accuracy is higher in hardware emulation than its software counterpart due to the undesirable current through the low efficacy and \textit{pruned} synapses. We further explore the performance of the multistate network for different network connectivity (\textit{C}) and input activity (\textit{f}). As shown in \fig{metric}-(f), at $\textit{C}=50\%$, the network mean accuracy drops with dense activity. When $\textit{C}=$ 40-60\%, the percentage of connected and \textit{pruned} synapses are comparable. Dense activity in this setting ($\textit{f}=$ 75-90\%) results in significant current through the \textit{pruned} synapses and the network shows poor performance which is observed in figure \fig{metric}(f). We deduce that the hardware multistate network can differentiate patterns well when there is a considerable difference between the number of connected and \textit{pruned} synapses, otherwise conflicting patterns and undesired current harshly affect the performance.
In order to demonstrate the functionality of the proposed scheme, a multistate synaptic network with 5 input neurons and 3 output neurons is simulated in Cadence Virtuoso. \fig{dual} (left panel) shows a case scenario where four synapses in a crossbar column, denoted by w$_1$-w$_4$, have active presynaptic inputs and the 
%>>>>> what is the error mean in this context? would not be better to say where the neuron output does not match the expected ground truth or label or .... Great....
% in the high level simulation section i specified which case is error 1 and -1, in one it would be potentiation and another depression so wanted to differentiate. I wonder if i should write equations. Like error= ground truth-network output. 
%> It is good to have some equations in the paper describing the learning rule. We can have underline the error term then
% okay i will add equations!
error is positive. According to the learning rule, a potentiating pulse is applied to these synapses. w$_2$,w$_3$ and w$_4$ have \textit{low} initial efficacy, but they are in metalevel 0, 1 and 2, respectively. 
%> which figure? you a little bit far 
%> Fatima, it would be good to make number the weight with subscript, what do you think? Okay... I will do it..
% yeah it would look better! Thank you!
%> Fatima, I would italicize the C, f, and other labels 
% on it
In \fig{dual}, we see that only w$_2$ changes its efficacy level due to potentiation, whereas w$_3$ and w$_4$ only transit to a lower metastate. w$_1$, which has \textit{high} initial efficacy at metalevel 0 transit to metalevel 1. The right panel of \fig{dual} shows the transitions of these potentiated weights when error is negative. Since w$_2$ is in metalevel 0, it changes its efficacy level to \textit{low} after the depression cycle while all the other weights retain their efficacy level with appropriate change in metastate. 
%Although emulation of metaplasticity has been investigated in device level in previous research, we have not found any work hardware realization of binary metaplastic synapses. This limits proper comparison of the presented results with other work.

%> Fatima, lets evaluate the power consumption in different subsection, what do you think?
% won't it be too short for a subsection?
%> This is not enough. It is too short. We need to talk about it more. Probably, you may want to describe the approach you used to estimate the power consumption. 
% i just averaged the power during the time of potentiation/depression
% I know.... It is up to you... by the end, it a suggestion. 
% I would like to write about it but I am afraid I do not have much to write. :( Had i more time I would have analyzed further
%> Quiting!! time to sleep... Rosl will kill me! 
% Thank you so much Abdullah!! 
%> You are welcome. ALL THE LUCK... GOOD NIGHT
% Gunnyt!!Thank you!!
% that should be synaptic efficacy...not just efficacy
%%%%%%%%%%%%%%%%%%%%%%%%%

\subsection{Power consumption}
% >>The power consumption "of" the proposed network will
% Suggestion: The power consumption of the proposed network is highly impacted by the input activity sparsity \textit{f} and network connectivity
The power consumption of the proposed network is highly impacted by the input activity (\textit{f}) and network connectivity (\textit{C}).
%>>>>> Suggestion: While considering the case scenario demonstrated in \fig{dual} where \textit{f}=0.75 and \textit{C}=0.5, we found out that the average power consumption during ...
 Considering connectivity \textit{C}=50\% in the implemented $5 \times 3$ crossbar network, we found the average power consumption to be $24.64\mu W$ (excluding the control circuitry)\footnote{Since we did not find any similar realization of binary metaplastic synapses, performance comparison could not be conducted.}for 100 input and output patterns with activity \textit{f}=75\%.
%Applying 100 input and output patterns  with activity \textit{f}=75\% and considering connectivity \textit{C}=50\% in the implemented $5 \times 3$ crossbar network, we found that the average power consumption is $24.64\mu W$ (excluding the control circuitry)\footnote{Since we did not find any similar realization of binary metaplastic synapses, proper performance comparison could not be drawn.}.
%While considering the case scenario demonstrated in \fig{dual} where \textit{f}=75\% and \textit{C}=50\%, we found that the average power consumption during potentiation is 26.98$\mu$W and it is 26.52$\mu$W during depression . 
%>>>> Suggestion: The higher power consumption during depression compared to potentiation attributes to the fact that the memristor has high initial conductivity so it draws higher current during transition
%The higher power consumption during depression compared to potentiation attributes to the fact that the memristor has high initial conductivity so it draws higher current during transition. 
Low input activity (\textit{f}) and low connectivity (\textit{C}) is highly favorable for the proposed network. In such setting, the power consumption of the network
%>>>>> Suggestion: replace 'a lot of' with the majority of the synaptic connections ...
is reduced since majority of the synaptic connections are pruned. It also enables the network to better utilize the metaplasticity of multistate synapses and show higher retention and learning capability.

%WHAT DOES THIS MEAN??? WHAT SHOULD WE INFER?

\section{Conclusions} 
Metaplastic synapses can equip neural networks to better address catastrophic forgetting. This work investigates the performance of multistate synapses for retention and reception of information. 
It is demonstrated that the model shows slower decay in the mean accuracy than binary model, with moderate deterioration in learning accuracy.  We then capture the characteristics of a multistate synapse in a memristive device through appropriate training method to map it to the metaplastic states. The inference and training procedure is validated by simulating a small scale crossbar network ($5\times3$ size) in Cadence.
Furthermore, high level emulation of the network shows that the number of patterns that the multistate memristive synaptic network can detect with $\leq$ 25\% mean error is $\simeq$ 2.1 times that of its binary counterpart. %Edge devices can benefit from employing a simple update in the synaptic device model without incurring additional resource or energy cost.
%\bibliographystyle{IEEEtran}
%\bibliography{IEEEabrv,refer}

\begin{thebibliography}{10}
\providecommand{\url}[1]{#1}
\csname url@samestyle\endcsname
\providecommand{\newblock}{\relax}
\providecommand{\bibinfo}[2]{#2}
\providecommand{\BIBentrySTDinterwordspacing}{\spaceskip=0pt\relax}
\providecommand{\BIBentryALTinterwordstretchfactor}{4}
\providecommand{\BIBentryALTinterwordspacing}{\spaceskip=\fontdimen2\font plus
\BIBentryALTinterwordstretchfactor\fontdimen3\font minus
  \fontdimen4\font\relax}
\providecommand{\BIBforeignlanguage}[2]{{%
\expandafter\ifx\csname l@#1\endcsname\relax
\typeout{** WARNING: IEEEtran.bst: No hyphenation pattern has been}%
\typeout{** loaded for the language `#1'. Using the pattern for}%
\typeout{** the default language instead.}%
\else
\language=\csname l@#1\endcsname
\fi
#2}}
\providecommand{\BIBdecl}{\relax}
\BIBdecl

\bibitem{abraham1996metaplasticity}
W.~C. Abraham and M.~F. Bear, ``Metaplasticity: the plasticity of synaptic
  plasticity,'' \emph{Trends in neurosciences}, vol.~19, no.~4, pp. 126--130,
  1996.

\bibitem{fusi2005cascade}
S.~Fusi, P.~J. Drew, and L.~F. Abbott, ``Cascade models of synaptically stored
  memories,'' \emph{Neuron}, vol.~45, no.~4, pp. 599--611, 2005.

\bibitem{leibold2007sparseness}
C.~Leibold and R.~Kempter, ``Sparseness constrains the prolongation of memory
  lifetime via synaptic metaplasticity,'' \emph{Cerebral Cortex}, vol.~18,
  no.~1, pp. 67--77, 2007.

\bibitem{zhu2017emulation}
X.~Zhu, C.~Du, Y.~Jeong, and W.~D. Lu, ``Emulation of synaptic metaplasticity
  in memristors,'' \emph{Nanoscale}, vol.~9, no.~1, pp. 45--51, 2017.

\bibitem{wu2018full}
Q.~Wu, H.~Wang, Q.~Luo, W.~Banerjee, J.~Cao, X.~Zhang, F.~Wu, Q.~Liu, L.~Li,
  and M.~Liu, ``Full imitation of synaptic metaplasticity based on memristor
  devices,'' \emph{Nanoscale}, vol.~10, no.~13, pp. 5875--5881, 2018.

\bibitem{kim2017nanogenerator}
B.-Y. Kim, H.-G. Hwang, J.-U. Woo, W.-H. Lee, T.-H. Lee, C.-Y. Kang, and
  S.~Nahm, ``Nanogenerator-induced synaptic plasticity and metaplasticity of
  bio-realistic artificial synapses,'' \emph{NPG Asia Materials}, vol.~9,
  no.~5, pp. e381--e381, 2017.

\bibitem{ben2007long}
D.~D. Ben Dayan~Rubin and S.~Fusi, ``Long memory lifetimes require complex
  synapses and limited sparseness,'' \emph{Frontiers in computational
  neuroscience}, vol.~1, p.~7, 2007.

\bibitem{jiang2017rram}
Y.~Jiang, J.~Kang, and X.~Wang, ``R{RAM}-based parallel computing architecture
  using k-nearest neighbor classification for pattern recognition,''
  \emph{Scientific reports}, vol.~7, p. 45233, 2017.

\bibitem{kvatinsky2015vteam}
S.~Kvatinsky, M.~Ramadan, E.~G. Friedman, and A.~Kolodny, ``V{TEAM}: A general
  model for voltage-controlled memristors,'' \emph{IEEE Transactions on
  Circuits and Systems II: Express Briefs}, vol.~62, no.~8, pp. 786--790, 2015.

\bibitem{zyarah2019neuromemrisitive}
A.~M. Zyarah and D.Kudithipudi, ``Neuromemrisitive architecture of {HTM} with
  on-device learning and neurogenesis,'' \emph{ACM Journal on Emerging
  Technologies in Computing Systems (JETC)}, vol.~15, no.~3, p.~24, 2019.

\bibitem{hu2011geometry}
M.~Hu, H.~Li, Y.~Chen, X.~Wang, and R.~E. Pino, ``Geometry variations analysis
  of tio 2 thin-film and spintronic memristors,'' in \emph{16th Asia and South
  Pacific Design Automation Conference (ASP-DAC 2011)}.\hskip 1em plus 0.5em
  minus 0.4em\relax IEEE, 2011, pp. 25--30.

\bibitem{zyarah2017ziksa}
A.~M. Zyarah, N.~Soures, L.~Hays, R.~B. Jacobs-Gedrim, S.~Agarwal,
  M.~Marinella, and D.~Kudithipudi, ``Ziksa: On-chip learning accelerator with
  memristor crossbars for multilevel neural networks,'' in \emph{2017 IEEE
  International Symposium on Circuits and Systems (ISCAS)}.\hskip 1em plus
  0.5em minus 0.4em\relax IEEE, 2017, pp. 1--4.

\end{thebibliography}


\end{document}